# Self-Supervised Borrowing Detection on Multilingual Wordlists

Tim Wientzek

University of Tübingen

Department of Linguistics

Keplerstr. 2, 72074 Tübingen, Germany

01.12.2025

**Abstract**

This paper presents a fully self-supervised approach to borrowing detection in multilingual wordlists. The method combines two sources of information: PMI similarities based on a global correspondence model and a lightweight contrastive component trained on phonetic feature vectors. It further includes an automatic procedure for selecting decision thresholds without requiring labeled data. Experiments on benchmark datasets show that PMI alone already improves over existing string similarity measures such as NED and SCA, and that the combined similarity performs on par with or better than supervised baselines. An ablation study highlights the importance of character encoding, temperature settings and augmentation strategies. The approach scales to datasets of different sizes, works without manual supervision and is provided with a command-line tool that allows researchers to conduct their own studies.

## 1 Introduction

Loanwords are among the most visible outcomes of language contact. They represent the transfer of lexical material from a donor language into a recipient language (Trask, 2000). Identifying and understanding borrowing is central to historical linguistics: on the one hand, filtering out loanwords allows to infer genealogical relationships between languages more accurately (List & Forkel, 2022; cf. Greenhill et al., 2009); on the other, their presence offers a window into past interactions, such as migrations, trade routes or patterns of cultural dominance (Montenegro et al., 2008; Haspelmath & Tadmor, 2009a; Miller, 2024).

To minimize the impact of borrowings, comparative studies often rely on Swadesh-style lists (Swadesh, 1955), which emphasize core vocabulary assumed to be diachronically stable. Yet even such lists are not immune to borrowing (Nelson-Sathi et al., 2011), and manual and especially automated identification remains difficult - especially for older, fully adapted loans that no longer reveal phonological traces of their donor origin. In practice, experts rely on a combination of phonological, morphological, and semantic evidence, as well as historical and areal context, to determine whether a word is likely to be borrowed rather than inherited (List, 2024). Automated models, in their current state, are not yet capable of integrating such diverse types of linguistic evidence. They are therefore typically constrained to identifying pairs of words across languages that are similar in both form and meaning, where common descent can be excluded.

Recent initiatives such as Lexibank (List et al., 2022) have provided large collections of standardized multilingual wordlists, facilitating data-driven exploration of lexical borrowing. Automatic borrowing detection allows rapid and systematic investigation of potential loanwords across language families or areas, serving as an exploratory step prior to detailed expert validation. Advances in automated borrowing detection therefore pave the way for large-scale reconstructions of contact networks, more robust phylogenetic inference, and quantitative models of areal influence.

Most existing computational approaches fall broadly into two categories. Phylogeny-based methods extend tree-based models to include horizontal transfer edges that capture borrowing events (e.g., Neureiter et al., 2022; Köllner, 2021). Sequence-comparison methods, in contrast, detect lexical similarities across languages using different distance metrics on concept-aligned wordlists (e.g., van der Ark et al., 2007; Zhang et al., 2021; List & Forkel, 2022; Miller & List, 2023). While more rare, approaches using only monolingual data have also been proposed, e.g. via detecting unusual phonotactic patterns (Kpoglu, 2025) that may indicate borrowing or training small language models (Miller et al., 2020). However, these methods often rely on specific preconditions, such as known phylogenies, labeled training data, or dominant donor languages (List, 2024; Alvarez-Mellado, 2025). Additionally, with only few exceptions such as the work by Neureiter et al. (2022), they typically cannot infer the direction of transfer. Moreover, they require similarity thresholds that are manually chosen (e.g. List & Forkel, 2022; Hantgan et al., 2022) or derived from training data (e.g. Zhang et al., 2021; Miller & List, 2023), which may limit generalization or lead to overfitting.

However, despite these advances, fully unsupervised or self-supervised methods for borrowing detection remain underexplored (List, 2024). This paper introduces a self-supervised approach to borrowing detection based on contrastive learning and probabilistic thresholding. The method learns similarity representations between phonetic sequences using SimCSE-style contrastive learning (Gao et al., 2022) combined with a pointwise mutual information (PMI) distance measure (Jäger, 2013). It further infers similarity thresholds automatically through Gaussian Mixture Models (cf. van der Ark, 2007), eliminating the need for labeled data or arbitrary parameter tuning.

The method is evaluated against the supervised approaches of Zhang et al. (2021) and Miller & List (2023), both of which employ 10-fold cross-validation using 90% of available data for training. The results show that the proposed self-supervised model combined with PMI similarity achieves comparable or superior performance without requiring supervision. Moreover, the evaluation suggests that PMI similarity alone provides a strong baseline for future experiments. These findings demonstrate the potential of self-supervised approaches for large-scale, exploratory borrowing detection across multilingual wordlists.

The remainder of this paper is structured as follows. §2 presents the data and methods, describing the datasets, data preparation, contrastive learning framework, model architecture, and evaluation procedure. §3 reports the results, followed by a discussion and outlook in §4.

## 2 Data & Methods

### 2.1 Datasets

All datasets used in this study are publicly available under a CC-BY 4.0 license. The primary dataset for model development and parameter selection was the one introduced by Miller & List (2023), which consists of seven South American language varieties that have Spanish as their dominant donor language (Table 1). The data originate from the *World Loanword Database* (WOLD; Haspelmath & Tadmor, 2009b) and the Intercontinental Dictionary Series (Key & Comrie, 2023). As in the original study, only borrowings from Spanish into these recipient varieties were considered to maintain comparability. Miller & List applied 10-fold cross-validation, stratified by concept, such that each word form appeared in the test data exactly once. The authors publicly released these data splits and code to replicate their supervised methods and run them on similar datasets. It should be noted that although data splits were available, the self-supervised model differs conceptually from the supervised ones: it does not rely on training/test partitions and instead uses 100% of the available data during representation learning.

Table 1. Overview of dataset from [Miller & List (2023)](#) with number of concepts, lexemes and loanword proportion in the complete set and the set with 75% of loanwords removed (reduced).

| Language | Complete | | | Reduced | | |
|---|---|---|---|---|---|---|
| | Concepts | Lexemes | Loanwords | Concepts | Lexemes | Loanwords |
| Imbabura Quechua | 1155 | 1156 | 26% | 929 | 929 | 9% |
| Mapudungun | 1040 | 1242 | 15% | 941 | 1083 | 4% |
| Otomi | 1252 | 2241 | 9% | 1205 | 2105 | 3% |
| Q'eqchi' | 1211 | 1773 | 9% | 1132 | 1649 | 3% |
| Wichí | 1128 | 1219 | 12% | 1028 | 1102 | 3% |
| Yaqui | 1242 | 1433 | 22% | 1062 | 1185 | 6% |
| Zinacantan Tzotzil | 955 | 1266 | 13% | 864 | 1131 | 3% |
| Spanish | 1308 | 1770 | - | 1308 | 1770 | - |
| Aggregate | 1308 | 12100 | 15% | 1308 | 12100 | 4% |

Their study evaluated several approaches, including normalized edit distance (NED; [Levenshtein, 1965](#); [Yujian & Bo, 2007](#)), sound-class-based alignment scores (SCA; [List, 2012](#)), and a linear support-vector machine classifier (SVM) trained jointly on both features. In [Miller (2024)](#), an additional feature based on "least cross-entropy" (LCE), the entropy of the product of conditional character probabilities from a small language model trained on the data, was introduced. The author could show that adding LCE as a feature to the classifier yielded the best overall results in their experiments on this dataset. These three configurations (NED, SCA, and the classifier trained on NED, SCA and LCE) were used as supervised baselines against which the self-supervised model was evaluated.

To determine the effects of dataset size and borrowing proportion on model robustness, two additional experimental settings were explored. First, to approximate data sizes typical for (computational) historical linguistic research, where wordlists often contain about 100 to 200 concepts (cf. Lexibank; [List et al., 2022](#)), compared to over 1000 per variety in WOLD, a reduced-data experiment was conducted. Each of the original test splits (10% of the data, i.e. about 120 concepts) was further divided into 10 folds, yielding training sets comprising 9% and test sets comprising 1% of the total data. Results were averaged across all 100 test splits.

Table 2. Overview of WOLD dataset (Haspelmath & Tadmor, 2009b) composed of English, Japanese and Thai.

| Language | Concepts | Lexemes | Loanwords |
|---|---|---|---|
| Thai | 1411 | 2096 | 2% |
| Japanese | 1449 | 2130 | 6% |
| English | 1411 | 1514 | - |
| Aggregate | 1456 | 5740 | 4% |

Table 3. Overview of dataset from Zhang et al. (2021), covering 49 Turkic and 39 Indo-Iranian varieties.

| Family | Concepts | Lexemes | Loanwords |
|---|---|---|---|
| Turkic | 183 | 2498 | 18% |
| Indo-Iranian | 183 | 1760 | 19% |
| Aggregate | 183 | 4258 | 18% |

Second, to evaluate the sensitivity of the models to different borrowing proportions and its cross-lingual generalization capabilities, two low-borrowing conditions were tested. For the first, 75% of the borrowings in the original Miller & List (2023) dataset were randomly removed, and new splits were created. This yielded a dataset with an average borrowing proportion of roughly 4% across languages (Table 1). For the second, this was done using the WOLD wordlists of English, Thai and Japanese, setting English as donor and Japanese and Thai as recipient languages (Table 2). On average, these two recipient languages have about 4% loanwords from English - much lower than the average of 15% loans from Spanish in the dataset used by Miller & List.

To further test how well the model generalizes cross-linguistically, it was also evaluated on the data used by Zhang et al. (2021). This dataset from Mennecier et al. (2016) contains 183 Swadesh-style concepts with up to 88 IPA-transcribed pronunciations each, covering 49 Turkic and 39 Indo-Iranian varieties (Table 3). The goal of that study was to detect potential borrowings between these families without specifying directionality. Pairwise similarity scores between all unique pronunciations were computed, and expert annotations were available for evaluation. The authors report results for SCA, PMI and spectrogram distance (Heeringa, 2004) in a 10-fold cross-validation setup, again using 90% of available data to determine the thresholds for classification and the remaining 10% for evaluation.

Together, these datasets and the accompanying studies provide complementary testing grounds and benchmarks to evaluate the model in settings with varying amounts of data, varying proportions of borrowings and typologically distinct language varieties.

## 2.2 Data preparation

All data used in this study consisted of concept-form mappings transcribed to the International Phonetic Alphabet (IPA). Following Wientzek (2025), using the Soundvectors package available in Python (Rubehn, 2024) the IPA strings were converted into sequences of phonetic feature vectors for the contrastive learning model, giving the model a more feature-rich initialization of the character embedding space. Each IPA character is represented by a 39-dimensional feature vector encoding place and manner of articulation, as well as features specific to diphthongs and tones. Each feature could take one of three values: 0 ("does not apply"), 1 ("not applicable"), or 2 ("applies"). This yields a compact and expressive representation in which most IPA characters are uniquely and meaningfully placed in feature space. An ablation study (§3.3) shows that using raw IPA symbols instead of feature vectors significantly reduces model performance.

In cases where an entry for a concept consisted of multiple words, usually separated by "+" or "_", the separators were removed, and the resulting sequence was treated as a single form. Apart from this, the IPA strings were preserved as is, including diacritics and symbols representing tones.

As in most previous work, potential borrowings were searched only among words referring to the same concepts. However, following Miller (2024), I also explored an extended evaluation that includes semantically related concepts based on colexification data from CLICS4 (Tjuka et al., 2025). Because not rigorously controlling the number of different concepts one considers drastically reduces performance, as already observed by Miller (2024), only concepts that colexified in at least 5% of the varieties in the CLICS database were considered. Additionally, similarities between words from different concepts were slightly penalized in the similarity score calculation by substracting 0.1 from the similarity score. Without such strict controls, the evaluation would suffer from too many false positives due to the increased probability of chance similarities between unrelated words from different concepts. As previously mentioned, the extension towards semantically related concepts should be regarded as purely exploratory. The 5% colexification threshold and the 0.1 similarity penalty were selected heuristically to balance coverage and noise rather than through systematic optimization. Their purpose is to evaluate the potential benefit of extending the borrowing detection to semantically related concepts rather than claiming empirical optimality.

## 2.3 Contrastive learning

Contrastive learning is a self-supervised approach that aims to structure the representation space such that similar items are close together and dissimilar items are far apart, all without requiring labeled data ([Hadsell et al., 2006](#)). It does so by contrasting positive pairs of inputs, two different, modified "views" of the same underlying item, with negative pairs drawn from other items in the same batch. The central idea is that the model should capture the essential characteristics of an item that stay consistent under small transformations or noise.

The specific approach adopted here builds on SimCLR ([Chen et al., 2020](#)) and SimCSE ([Gao et al., 2022](#)), which introduced self-supervised methods for learning from modified views of images and sentences, respectively. In SimCLR, each image in a batch is transformed twice through random augmentations such as cropping, adding noise, color distortion, or rotation. These two transformed versions form a positive pair, while all other images (and their augmentations) from the same batch act as negative examples. Both views of each image are passed through an encoder to obtain hidden representations, and then through a projection head, which is typically a small multilayer perceptron followed by a nonlinear activation (see Figure 1). The model learns by using a contrastive loss (see [§2.4](#)) that maximizes the similarity between projections of the positive pairs while minimizing similarity with all negatives. Through this process, the model learns to form a latent space in which the hidden representations of semantically similar inputs cluster together. SimCSE applied this procedure to sentences and learning sentence embeddings. Notably, [Chen et al. (2020)](#) and [Gao et al. (2022)](#) showed that training on the projections rather than directly on the hidden representations improves the overall quality of the learned representations.

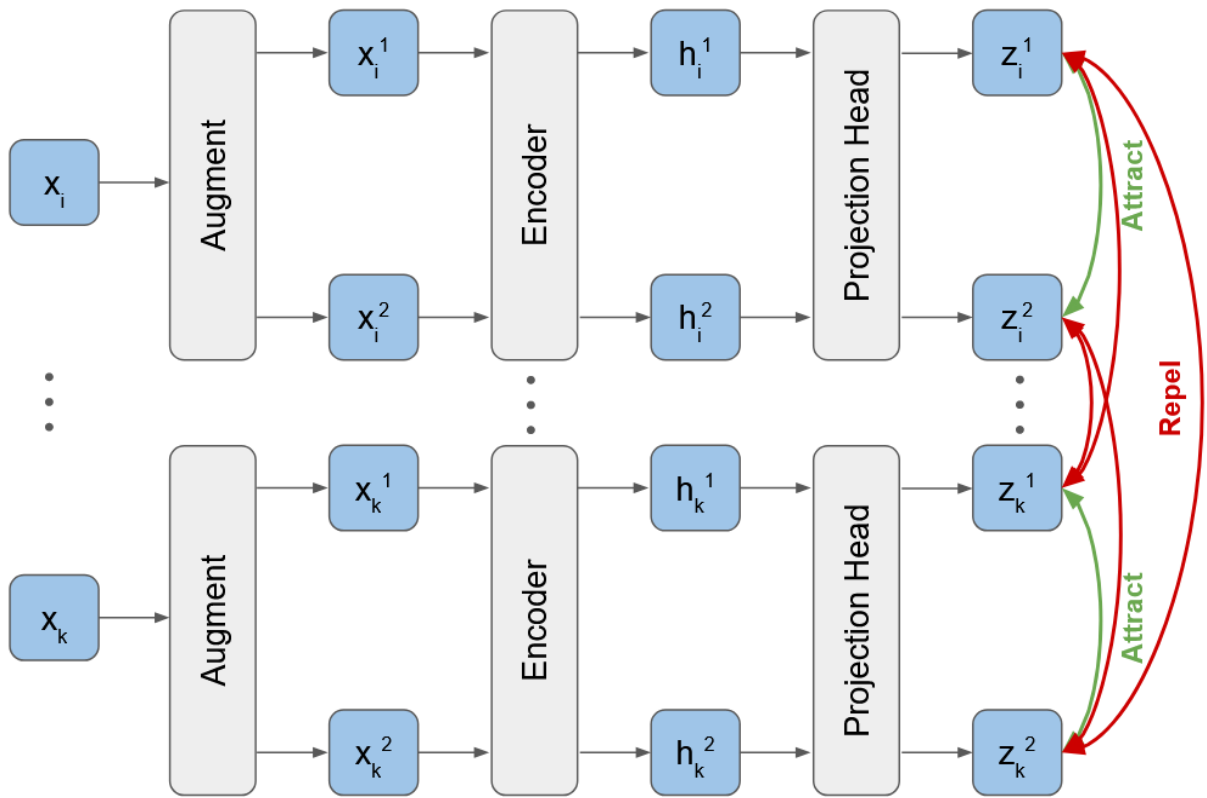

Figure 1: Model overview, adapted from Chen et al. (2020). Each batch item x is passed through the model twice to get two augmented views, which are then processed by the encoder, pooled and passed through a projection head. The model objective (loss function) maximizes agreement between the two projected augmented views of an item (attract), while minimizing agreement between these and the projections from all other items (repel).

Applied to the setting of borrowing detection, the same principle is transferred to phonetic word forms, where each form is represented as a sequence of phonetic feature vectors (see §2.2). Borrowed words often exhibit patterns that resemble the "augmented" views used in contrastive frameworks: they preserve parts of the phonetic structure of their donor form while being modified to match the phonology and phonotactics of the recipient language. By training the model to pull together the noisy variants created via augmentation, it may capture some degrees of phonetic and structural similarity in its learned representations. This contrastive framework thus allows the model to learn a representation space in which phonetically similar word forms, such as potential borrowings, should cluster together, without requiring any labeled training data.

The next section details the model architecture and training setup implementing this framework.

## 2.4 Model and training procedure

The model takes padded sequences of phonetic feature vectors representing word forms as input (see §2.2) and projects them into a higher-dimensional embedding space using a linear layer, followed by layer normalization and feature dropout, similar to the procedure in SimCSE.

Following Wu et al. (2022; ESimCSE), each word can have one of its characters duplicated with a set probability. In their paper, the authors present a slightly improved version of SimCSE and show that, because positive pairs consist of sentences of equal length, cosine similarity tends to be strongly correlated with sentence length. This effect could be reduced by duplicating words within the sentence, whereas just inserting or deleting random words had a negative effect.

Additionally, Gaussian noise is added to the word embeddings with a set probability as part of the augmentation strategy. Sinusoidal positional encodings are then added to retain sequential information (Vaswani et al., 2017). The resulting sequence is passed through a Transformer Encoder (Vaswani et al., 2017), with the addition of relative positional encodings as implemented in the x-transformers library (Su et al., 2023; Wang, 2023). Standard padding masking is applied. From the final encoder layer, representations are aggregated by averaging over the valid positions. For training, a projection head is applied to the pooled representations to compute the contrastive loss; during evaluation, only the hidden representations from the encoder are used, in line with SimCLR and SimCSE. Figure 1 provides a schematic overview of the model architecture.

During training, each word is presented to the model twice, creating two slightly different augmented views. Because the augmentations involve random character duplication, Gaussian noise, and dropout in the Transformer, the augmented views vary across epochs. Mini-batches are constructed in a way that forms for a concept are evenly distributed across batches, minimizing the likelihood of the same concept appearing multiple times within a batch, reducing the risk of treating potential borrowings as negative pairs.

While stronger augmentations as used in SimCLR may be effective for image data, previous work on sentence embeddings (SimCSE and ESimCSE) found that only standard Transformer dropout (0.1) and character duplication were superior to more complex transformations. In the current, character-based setting, adding Gaussian noise to the feature embeddings further helps the model account for phonetic adaptations in borrowings, which can be viewed as noisy variants of the donor word.

Once the augmented views are obtained and passed through the encoder and projection head, the model is trained using a contrastive loss that encourages the projections of views of the same word to be similar while pushing apart projections of different words. Formally, for each word $x_i$, we denote its two augmented views as $x_i^1 = a(x_i)$ and $x_i^2 = a(x_i)$, with corresponding hidden representations $h_i^1 = f(x_i^1)$ and $h_i^2 = f(x_i^2)$ and projections $z_i^1 = g(h_i^1)$ and $z_i^2 = g(h_i^2)$, where $a(\cdot)$, $f(\cdot)$, and $g(\cdot)$ are the augmentation module, encoder, and projection head, respectively (see Figure 1).

The loss of a positive pair of projections of $x_i$ is calculated as

$$\ell_{i^1,i^2} = -\log \frac{e^{\frac{sim(z_i^1, z_i^2)}{\tau}}}{\sum_{j=1}^{2N} \mathbb{1}_{[i \neq j]} e^{\frac{sim(z_i, z_j)}{\tau}}}, \qquad (1)$$

where $sim(\cdot)$ is the cosine similarity, which is equivalent to the dot product after l2-normalization, i.e.

$$sim(u, v) = \frac{u^T v}{\|u\| \cdot \|v\|}. \qquad (2)$$

$\mathbb{1}_{[i \neq j]}$ is an indicator function that evaluates to 0 if two views belong to the same items and 1 otherwise and $\tau$ is the temperature, a hyperparameter that regulates how sharp the distinctions between positive and negative samples should be treated. A lower temperature leads to sharper contrast while a higher temperature leads to smoother contrast.

The model was trained for four epochs using the PyTorch ([Paszke et al., 2019](#)) implementation of the AdamW optimizer ([Loshchilov & Hutter, 2019](#)), with parameters listed in Table 4. These settings were determined through manual experimentation on the [Miller & List (2023)](#) dataset rather than through extensive hyperparameter optimization. Due to aforementioned sampling strategy, the batch size was selected such that forms for each concept were roughly evenly distributed across batches. As a practical guideline, the number of batches should exceed the number of forms per concept (approximately equal to the number of languages in the dataset), while still aiming for reasonably large batch sizes (see §3.3). Temperature and dropout values follow recommendations from the literature (Gao et al., 2022; Wu et al., 2022), and the learning rate was set to the AdamW default. With these settings, the model is lightweight and fast, containing approximately 865k parameters and completing a full training run on the dataset by Miller & List (2023) in under 20 seconds on a single NVIDIA 3070 Laptop GPU.

Table 4. Default settings of the hyperparameters and augmentation probabilities used for all experiments.

| Parameter | Value |
| --- | --- |
| Hidden Dimension | 256 |
| Encoder Layers | 1 |
| Encoder Heads | 4 |
| Feature Dropout | 0.1 |
| Attention Dropout | 0.1 |
| Noise Probability | 0.5 |
| Duplication Probability | 0.1 |
| Temperature | 0.05 |
| Learning Rate | 0.001 |
| Batch size | 128 |

## 2.5 Evaluation

For inference, the original (non-augmented) word forms were passed through the model encoder to obtain their hidden representations. For each word, similarity scores were then computed against all other words referring to the same concept but belonging to different language families. The overall similarity between two words was defined as a weighted combination of the cosine similarity between their L2-normalized hidden representations and the pairwise PMI similarity.

PMI similarities were computed using the pre-computed scores for the 41 characters of the Automated Similarity Judgment Program transcription system (ASJP; Wichmann et al., 2025; Brown et al., 2013) made publicly available by Jäger (2018). PMI quantifies how strongly two sounds tend to co-occur in cognates beyond what would be expected by chance, with higher values indicating stronger associations. The PMI similarity between two words is the sum of the correspondence scores in their pairwise alignment (see Jäger, 2013, for details) and was min-max scaled to values between 0 and 1 here. Since the PMI scores are defined over the ASJP transcription system, all IPA strings were automatically converted to ASJP using the built-in functions of the LingPy library (List & Forkel, 2021) before calculating the similarities.

The combination of cosine and PMI similarities was determined empirically on the dataset by Miller & List (2023), resulting in weights of 25 % for cosine similarity and 75 % for PMI similarity. This weighting was kept fixed for evaluations on all datasets to maintain the unsupervised nature of the approach, but it should be noted that, while giving robust results, it is very likely not to be optimal combination for most datasets.

For classification, a borrowing situation was assumed when the highest similarity score of a candidate word exceeded a pre-defined threshold (see §2.6). Thresholds were determined separately for each recipient language, although determining a single global threshold may be preferable when less data is available.

The metrics used for evaluation were Precision, Recall, F1-Score and Accuracy, as done in the studies that served as baselines to the method presented here. To ensure comparability, the evaluation protocols of the respective studies were replicated. In Miller & List (2023), only borrowings from the set donor language were considered, and a prediction was counted as correct only if both the borrowing status and donor source were identified correctly. By contrast, Zhang et al. (2021) evaluated over all pairwise combinations, i.e. if a pair was labeled and predicted as borrowing, it counted as true positive. Because their dataset contains multiple near-identical pronunciations per concept, identifying only the closest candidate, as done in the other setting, would yield different outcomes from their original setup.

The results section also contains several ablation experiments, where varying degrees of probabilities for the augmentations (noise, duplication and dropout), adding swapping and deletions as augmentations, different temperatures for the loss function, removing the projection head, using raw IPA characters instead of phonetic feature vectors and increased numbers of encoder layers were tested.

## 2.6 Threshold selection

The main source of supervision in previous approaches to automatic borrowing detection via sequence comparison lies in the choice of a threshold for classification (van der Ark et al., 2007; Zhang et al., 2021; List & Forkel, 2022; Miller & List, 2023). All such methods compute pairwise similarity scores between words from two different languages and then, based on a threshold that is either predefined or derived from training data, classify each pair as containing a borrowing or not. Therefore, for a fully self-supervised approach, the threshold must instead be determined automatically in a data-driven way.

To achieve this, all similarity scores for a given recipient language are aggregated, and two Gaussian Mixture Models (GMMs) are fitted to the resulting distribution: one with a single component and one with two components (see van der Ark et al., 2007). GMMs assume that the observed distribution is generated from a mixture of Gaussian distributions. For both models, the Bayesian Information Criterion (BIC) is computed, and the model with the lower BIC is selected as the better fit.

If the two-component model fits better, this suggests that the similarity distribution is bi-modal, which would be expected if the data contain both unrelated word pairs (clustered around low similarity values) and pairs related via borrowing or cognacy (clustered around higher similarity values). The classification threshold can then be defined as the intersection point between the two Gaussian components, i.e. the similarity score at which both components contribute equally to the mixture. Formally, for a mixture $p(x) = f_1(x) + f_2(x)$, the intersection (i.e. the threshold) is the value where $f_1(x^*) = f_2(x^*)$. This can be analytically solved using Brent's method to find the root for $g(x) = f_1(x) - f_2(x)$ (Brent, 1973).

In cases where the single-component GMM provides a better fit, for example because there are not enough borrowings in the data to detect a distinct cluster, a simple fallback strategy is applied. The threshold is then set to one standard deviation above the mean of the similarity distribution, corresponding to the tail region where potential borrowings are most likely located. It should be noted that this threshold only offers a rough separation of the similarity scores. When the fallback strategy is applied, examining the similarity distribution manually is recommended. This can be carried out using the visualization features of the accompanying command-line tool and then setting the threshold(s) manually.

## 3 Results

### 3.1 Main results

This section presents the results for the experimental settings described in §2. For all settings, averages and standard deviations over five runs are reported for the methods involving contrastive learning, reflecting the non-deterministic nature of the training process and augmentation procedure.

Table 5. Results on the complete dataset from Miller & List (2023). For the methods involving contrastive learning, the values are given as the mean and standard deviation (in brackets) across 5 independent runs.

| Method | Supervision | Precision | Recall | F1 | Accuracy |
|---|---|---|---|---|---|
| NED (Miller & List, 2023) | Yes | 0.832 | 0.703 | 0.761 | 0.938 |
| SCA (Miller & List, 2023) | Yes | 0.869 | 0.72 | 0.787 | 0.945 |
| Classifier (NED, SCA, LCE) (Miller, 2024) | Yes | 0.871 | 0.827 | 0.848 | 0.958 |
| NED | No | 0.809 | 0.71 | 0.756 | 0.935 |
| SCA | No | 0.828 | 0.738 | 0.781 | 0.941 |
| Cosine similarity | No | 0.910 (0.01) | 0.645 (0.011) | 0.755 (0.006) | 0.940 (0.001) |
| PMI | No | 0.913 | 0.754 | 0.826 | 0.955 |
| Combined | No | 0.909 (0.002) | 0.769 (0.002) | 0.833 (0.002) | 0.956 (<0.001) |
| Combined + similar concepts | No | 0.913 (0.004) | 0.793 (0.002) | 0.849 (0.001) | 0.960 (<0.001) |

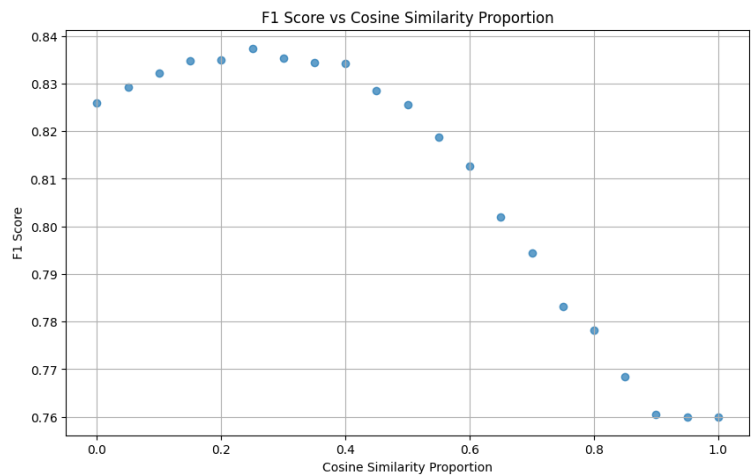

Figure 2: Effect of cosine similarity proportion in the weighted similarity scoring on the F1 score.

Table 5 shows the results on the full dataset from Miller & List (2023), together with the main scores reported in their work and in Miller (2024). PMI similarity performs strongly on its own, and combining cosine similarity from the contrastively learned representations with PMI yields a slightly higher F1 score. In contrast, using only cosine similarity leads to a significantly lower recall and a F1 score close to the NED baseline. To allow a direct comparison of thresholding strategies, results for NED and SCA with automatic thresholding are also provided. Here, the scores are slightly below, but comparable to the supervised thresholds. With the addition of similar concepts, i.e. concepts that commonly colexify across languages, the combined model reaches an F1 score comparable to the best result reported by Miller (2024).

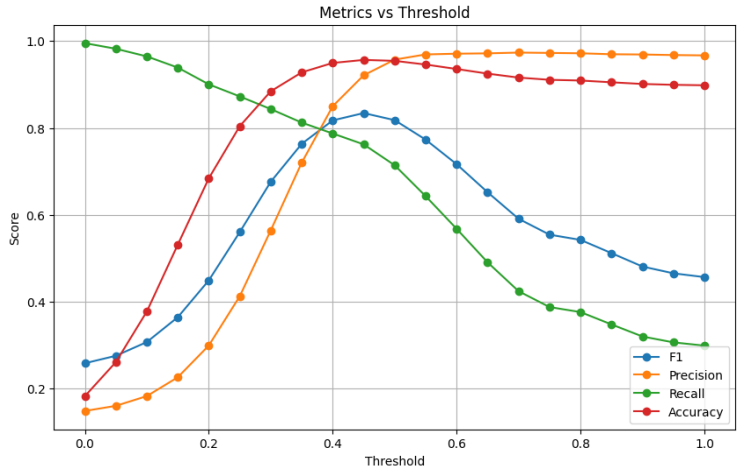

*Figure 3: Effect of global thresholds on precision, recall, F1 score and accuracy. Higher values indicate a higher minimum similarity of a word pair to be classified as borrowing.*

Figure 2 shows how the F1 score changes when varying the weight of cosine similarity in the combined model. For this dataset, weights between roughly 5-45% improve performance compared to PMI alone, with the best configuration at about 25%.

To examine the effect of threshold selection, metrics were also calculated using manually set global thresholds (Figure 3). Unsurprisingly, lowering the threshold, and therefore classifying more candidates as borrowings, increases recall and decreases precision and vice versa. Therefore, if, for example, one wishes to find more potential borrowings, although at the cost of more false positives, they may choose to slightly decrease the threshold compared to the one selected by the model.

*Table 6. Results on the dataset from Miller & List (2023), where only 10% of data was available. For the methods involving contrastive learning, the values are given as the mean and standard deviation (in brackets) the 10 folds. For the supervised methods, the results represent averages over the 10-fold cross-validation over the 10 folds.*

| Method | Supervision | Precision | Recall | F1 | Accuracy |
|---|---|---|---|---|---|
| NED (Miller & List, 2023) | Yes | 0.854 | 0.687 | 0.743 | 0.938 |
| SCA (Miller & List, 2023) | Yes | 0.831 | 0.707 | 0.75 | 0.94 |
| Classifier (NED, SCA, LCE) (Miller, 2024) | Yes | 0.721 | 0.552 | 0.605 | 0.905 |
| Cosine similarity | No | 0.781 (0.101) | 0.669 (0.035) | 0.718 (0.048) | 0.926 (0.013) |
| PMI | No | 0.89 | 0.746 | 0.81 | 0.951 |
| Combined | No | 0.912 (0.027) | 0.728 (0.045) | 0.808 (0.023) | 0.951 (0.008) |
| Combined + similar concepts | No | 0.915 (0.026) | 0.732 (0.043) | 0.812 (0.022) | 0.952 (0.007) |

Table 7. Results on the dataset from Miller & List (2023), where only 75% of borrowings were removed. For the methods involving contrastive learning, the values are given as the mean and standard deviation (in brackets) over 5 independent runs.

| Method | Supervision | Precision | Recall | F1 | Accuracy |
|---|---|---|---|---|---|
| NED (Miller & List, 2023) | Yes | 0.828 | 0.629 | 0.692 | 0.977 |
| SCA (Miller & List, 2023) | Yes | 0.831 | 0.707 | 0.71 | 0.94 |
| Classifier (NED, SCA, LCE) (Miller, 2024) | Yes | 0.764 | 0.589 | 0.658 | 0.975 |
| Cosine similarity | No | 0.795 (0.028) | 0.625 (0.011) | 0.699 (0.01) | 0.978 (0.001) |
| PMI | No | 0.811 | 0.724 | 0.765 | 0.982 |
| Combined | No | 0.830 (0.008) | 0.726 (0.011) | 0.774 (0.004) | 0.982 (<0.001) |
| Combined + similar concepts | No | 0.831 (0.013) | 0.746 (0.007) | 0.786 (0.003) | 0.983 (<0.001) |

Table 8. Results on the WOLD dataset (Haspelmath & Tadmor, 2009b) composed of English, Japanese and Thai. For the methods involving contrastive learning, the values are given as the mean and standard deviation (in brackets) over 5 independent runs.

| Method | Supervision | Precision | Recall | F1 | Accuracy |
|---|---|---|---|---|---|
| NED (Miller & List, 2023) | Yes | 0.542 | 0.631 | 0.575 | 0.966 |
| SCA (Miller & List, 2023) | Yes | 0.59 | 0.727 | 0.644 | 0.969 |
| Classifier (NED, SCA, LCE) (Miller, 2024) | Yes | 0.787 | 0.471 | 0.581 | 0.975 |
| Cosine similarity | No | 0.744 (0.031) | 0.600 (0.060) | 0.662 (0.030) | 0.977 (0.001) |
| PMI | No | 0.715 | 0.842 | 0.773 | 0.981 |
| Combined | No | 0.717 (0.017) | 0.859 (0.007) | 0.782 (0.011) | 0.982 (0.001) |
| Combined + similar concepts | No | 0.732 (0.014) | 0.837 (0.003) | 0.780 (0.007) | 0.982 (<0.001) |

Table 9. Results on the dataset from Zhang et al. (2021). For the methods involving contrastive learning, the values are given as the mean and standard deviation (in brackets) over 5 independent runs. Accuracy was not measured in Zhang et al. (2021) and is therefore also omitted here.

| Method | Supervision | Precision | Recall | F1 |
|---|---|---|---|---|
| PMI (Zhang et al., 2021) | Yes | 0.742 | 0.813 | 0.764 |
| SCA (Zhang et al., 2021) | Yes | 0.863 | 0.815 | 0.821 |
| Cosine similarity | No | 0.676 (0.025) | 0.745 (0.019) | 0.708 (0.007) |
| PMI | No | 0.809 | 0.829 | 0.819 |
| Combined | No | 0.835 (0.005) | 0.816 (0.005) | 0.826 (0.003) |

The effects of automated thresholding become clearer in the settings where less data was available (Table 6) or where the borrowing proportion was substantially reduced or lower due to different languages in the data (Table 7 and Table 8). In these settings, the supervised models - especially the classifier - are outperformed by both PMI similarity and the combined model. As mentioned earlier, all model hyperparameters were kept constant across experiments and were not re-tuned for these settings. Overall, these results suggest that adding cosine similarity from the learned representations to PMI usually produces results that are at least on par with PMI similarity alone, with the most consistent gains appearing in precision.

Table 9 reports the results on the dataset used by Zhang et al. (2021). The scores were recalculated because several F1 values in the original publication's table were incorrect (cf. Table 5 in the original study). Since the source code and the paper lack information on how ten-fold cross-validation was performed, it was not possible to replicate the exact results reported in the study. Except for data splitting, the original code was used to calculate the scores based on the distances provided with the paper. Most notably, the PMI scores obtained from a global dataset on ASJP characters instead of scores calculated for IPA characters and only using the data at hand yield significantly higher performance than those reported in the original study.

## 3.2 Qualitative analysis

To better understand why combining cosine similarity from the learned representations with PMI similarity often outperforms either method individually, it is useful to examine the differences in their predictions. Using the results for the Miller & List (2023) dataset as an example, we can inspect word pairs that were predicted to be borrowings by one method but missed by another (Table 10).

Pairs detected by cosine similarity but not by PMI similarity include cases such as *aɲara* <-> *araɲa*, where segment order differs. Because the contrastive model produces pooled sequence representations, it can be less sensitive to processes such as metathesis that negatively impact alignment-based methods like PMI. Another factor is that PMI similarity is computed as the sum of correspondence scores from the alignment. Very short words therefore receive low PMI values even when they match closely, whereas cosine similarity is indifferent about absolute word length (cf. below).

PMI similarity, on the other hand, detects many pairs missed when only using the learned representations. These are typically cases where the words align cleanly and contain highly informative correspondences, even when insertions or deletions occur. The combined model captures additional cases that are missed by PMI similarity alone. These may be pairs that show good surface similarity but contain characters with low PMI informativeness (e.g. *kanan <-> ganar*), where the added high cosine similarity pushes the combined score over the threshold. Surprisingly, while the hidden representations of pairs like *rueram <-> rweða* seem to be closer in embedding space, some more obvious similarities such as *wela <-> bela* are missed by cosine similarity. A small set of borrowings is recovered only by the combined model but by neither individual similarity score. These cases generally involve moderate alignment quality together with moderate representational similarity, where neither signal is strong enough alone, but their weighted combination barely surpasses the decision threshold.

Table 10. Words in the Latin American language varieties from the complete Miller & List (2023) dataset correctly identified as borrowings from Spanish (words in brackets) by given similarity measure. Numbers in brackets indicate the total number of borrowings uniquely identified when using one method in contrast to the other. These values contain overlaps, e.g. borrowings detected by PMI, but not by cosine may also be included in borrowings detected by the combined measure, but not PMI.

| Detected by cosine, but not by PMI (22) | Detected by PMI, but not by cosine (145) | Detected by combined, but not PMI (27) | Detected by combined, but not cosine (148) | Detected by combined, but neither by PMI nor cosine (8) |
|---|---|---|---|---|
| awʒana (auɾar) | adubi (aðoβe) | indindina (entendɛr) | wela (bela) | ajuːninkriʃ (aɟunar) |
| o (o) | θepwe (dɛspwes) | sindina (sentir) | fazofe (aðoβe) | leʔiaroa (leɛr) |
| rueram (rweða) | ʔalkuʔ (arko) | boi (bwei) | kuka (kukaratʃa) | zefen (deβɛr) |
| ofaz (uβa) | kalpintelu (karpintɛro) | poːs (fosforo) | nʷʰebe (xweβes) | fɨlaŋ (blanko) |
| aɲara (araɲa) | lagrio (laðriɟo) | kanan (ganar) | reʃtiko (tɛstiɣo) | ʃeːr (sjɛra) |

Table 11. Mean and variance of word length differences in word pairs for borrowings uniquely detected by either measure. A higher variance given a similar mean indicate that the measure detects borrowings in word pairs with larger word length differences.

| Method | Mean | Variance |
|---|---|---|
| PMI | 0.69 | 0.738 |
| Cosine | 0.636 | 0.322 |

Table 11 shows the mean and the variance of word lengths in pairs exclusively detected by either similarity score. It seems that PMI is more tolerant of larger length differences as long as partial alignments are informative (such as *θepwe <-> dɛspwes*; see Table 10), while cosine similarity is more sensitive to substantial differences in lengths (cf.

Wu et al., 2022). In theory, this could be improved by introducing more or stronger augmentations that change the word lengths, such as more duplication, deletion or cutting parts of the word. However, these did not lead to improvements in the classification task (see §3.3) and require further investigation.

As already noted by Miller & List (2023), some false positives by either method can be retraced back to cases of chance similarity (e.g. due to probable onomatopoeia such as in Spanish *beβe* "BABY" vs. Q'eqchi *baɓaj*). For others, erroneous annotation of the underlying data is more likely, such as Spanish *pelota* "BALL" vs. Wichi *pelutax*, marked as non-borrowed in the data. It is reasonable to assume that the resulting detection errors occur to a similar degree in most methods and should therefore not impact comparability.

## 3.3 Ablation study

The ablation study (Table 12) reports F1 scores averaged over five independent runs for settings in which a single parameter was modified relative to the default configuration (Table 4). The results show that performance is most sensitive to a small number of factors. The strongest effect comes from data encoding: replacing phonetic feature vectors with raw IPA symbols leads to one of the largest performance drops in both the contrastive and combined models, indicating that the feature representation contributes substantially to learning useful similarities in data-sparse conditions (cf. Wientzek, 2025).

The temperature parameter, where higher values lead to flatter similarity distributions, is also highly impactful. While slightly higher values (e.g., 0.1 vs. 0.05 in the default configuration) can yield marginal improvements, performance degrades rapidly as temperature increases (cf. Gao et al., 2022). Adding deletions as an additional augmentation technique consistently reduces performance across all tested probabilities. Even though deletion and swapping are only applied to words with at least four segments to avoid overly disrupting short word forms, deletions still remove too much structural information for the model to benefit (cf. Gao et al., 2022). By contrast, swapping adjacent characters has virtually no effect in any configuration, suggesting that the model is either robust to or largely unaffected by such a reordering operation.

Table 12. Effects of different parameter and augmentation probability settings on the evaluation when using only the learned representations (cosine) or the combined similarity score. **Bold** values are significant with p-value < 0.05 (paired t-test).

| Factor | Setting | Cosine | Combined |
|---|---|---|---|
| - | Original | *0.755* | *0.834* |
| Projection head | removed | **0.734** | 0.831 |
| Data encoding | IPA | **0.647** | **0.815** |
| Batch sampling | Random | 0.753 | 0.834 |
| Temperature | 0.1 | 0.764 | 0.830 |
| | 0.25 | **0.740** | **0.820** |
| | 0.5 | **0.696** | **0.763** |
| | 1 | **0.605** | **0.677** |
| Batch size | 64 | **0.739** | 0.833 |
| | 256 | 0.757 | 0.833 |
| Number of layers | 2 | **0.744** | 0.832 |
| | 4 | **0.716** | **0.827** |
| Noise | 0.0 | 0.75 | 0.834 |
| | 0.25 | 0.758 | 0.833 |
| | 0.75 | 0.757 | 0.834 |
| Duplication | 0.0 | **0.729** | 0.833 |
| | 0.25 | 0.755 | 0.834 |
| | 0.5 | 0.752 | 0.834 |
| | 0.75 | 0.756 | 0.834 |
| Swapping | 0.1 | 0.756 | 0.833 |
| | 0.25 | 0.757 | 0.833 |
| | 0.5 | 0.753 | 0.834 |
| | 0.75 | 0.758 | 0.833 |
| Deletion | 0.1 | 0.748 | **0.821** |
| | 0.25 | **0.740** | **0.820** |
| | 0.5 | **0.735** | **0.808** |
| | 0.75 | **0.731** | **0.827** |
| Attention dropout | 0.0 | 0.75 | 0.833 |
| | 0.2 | 0.753 | 0.834 |
| | 0.3 | 0.753 | 0.833 |
| Feature dropout | 0.0 | **0.747** | 0.833 |
| | 0.2 | 0.75 | 0.833 |
| | 0.3 | 0.752 | 0.834 |
| Only dropout (both types) | 0.1 | **0.724** | 0.832 |

Removing duplication, increasing the number of layers, decreasing the batch size or removing the projection head mostly negatively affect the performance of the contrastive model, while the PMI similarity in the combined model partially compensates for the weaker learned representations. Sampling batches at random and the remaining settings do not noticeably affect the scores of either method for this specific dataset, indicating that the model is relatively robust to these choices.

Overall, the ablation results highlight that the main drivers of performance are the representational encoding, the temperature of the contrastive objective, and the avoidance of structurally too destructive augmentations.

## 4 Discussion & Outlook

The present work advances borrowing detection through a self-supervised framework that does not rely on any labeled data. Its contributions are threefold: application of robust PMI similarity grounded in global correspondence scores, a lightweight contrastive component that learns phonetic representations directly from the data, and an automatic thresholding procedure that removes the need for selecting thresholds manually or based on training data. Taken together, these elements yield a method that performs on par or outperforms supervised baselines in my experiments while remaining simple, fast to train and scalable across dataset sizes and borrowing proportions. The accompanying code provides options to choose (or not choose) the donor language, adjust the weighting between PMI and cosine similarity, modify all relevant parameters, display the similarity distributions, or experiment with manual thresholds via the command line.

A central result is that automatic threshold selection performs surprisingly well (cf. [van der Ark et al., 2007](#)). For both NED and SCA, the inferred thresholds closely match those derived from fully supervised training, indicating that the distributional shape of the similarity scores themselves often contains sufficient information to delimit likely borrowings. This enables the application of borrowing detection methods in realistic scenarios with sparse or missing borrowing annotations. In datasets with very low borrowing proportions, the model may employ a fallback (mean plus standard deviation) if no clear bi-modal distribution is present. In such cases, manually setting a threshold, informed by visual inspection of similarity score distributions for example, is recommended for researchers intending to conduct their own studies. For the sake of maintaining a fully self-supervised evaluation, this was not applied in the experiments presented here. An additional benefit of automatic thresholding, and of the self-supervised setup more generally, is that the model does not overfit threshold decisions to the distribution of an annotated training set, as may happen in supervised classifiers. As the experiments show, this contributes to its robustness across different borrowing proportions and dataset sizes.

The strength of PMI similarity stands out clearly in the results. Even without learned representations, PMI alone yields substantial improvements over previously proposed applications and over other phonetic string similarity measures such as NED and SCA. Part of this contrast can be attributed to methodological weaknesses in earlier work. For example, Zhang et al. (2021) relied on IPA-based PMI scores derived directly from their dataset, i.e. over a very fine-grained character inventory, which were likely too sparse to generalize well. By contrast, computing PMI scores over a more compact inventory such as the ASJP encoding produces stable cross-linguistic correspondences (cf. Jäger et al., 2017; Jäger, 2018). This stability and cross-lingual applicability make PMI a strong baseline in its own right.

The contrastive model, on the other hand, contributes a different but complementary type of information. Its advantages become most visible in cases where the PMI signal is weakened, for instance when correspondences are uninformative or words do not align well. Because the model operates on pooled feature-vector representations, it is less sensitive to ordering differences than alignment-based measures, and the feature vectors offer a fine-grained initial character representation. At the same time, its behaviour reveals interesting patterns that deserve further investigation. For example, occasional false negatives on highly similar surface forms, such as *wela <-> bela*, suggest a greater influence of word-initial phonemes on the similarity of learned representations in the embedding space. This could reflect positional biases found in Transformer-based architectures, as suggested by Wu et al. (2025), inter alia, who identify a strong bias towards first positions in sequences in attention mechanisms.

While PMI similarity captures systematic phonetic correspondences, it scales with word length, therefore penalizing short forms, and depends on alignment structure. Contrarily, cosine similarity is length-invariant (for pairs of comparable length) and captures more general phonetic resemblance independent of explicit segment matching. Combining both measures allows each to compensate for the other's weaknesses. As a result, the combined model identifies word pairs that align cleanly and exhibit informative correspondences, while also detecting those that share phonetic similarity despite not properly aligning or are of very short length. This balance leads to consistently strong performance across dataset sizes, languages and borrowing proportions.

Experiments on including semantically close concepts when searching for borrowings show promising results, uncovering additional borrowings (cf. Miller, 2024). As mentioned earlier, this requires carefully restricting the search space to reduce the risk of introducing false positives due to the increased likelihood of chance similarities associated with a broader search. Additionally, benefits of broadening the search require a sufficiently rich concept set, as classical Swadesh-style lists likely contain less concepts that commonly colexify.

The ablation studies highlight the importance of certain parameters for detection performance. Most notably, the model benefits from encoding the input as feature vectors instead of using raw IPA characters. Additionally, low temperatures yield the best results, which may also explain why even small amounts of deletion harm performance (see Table 12). Lower temperatures result in sharper distinctions in representation space, and phonemes removed entirely from a form affect the structure and thus the pooled representation more than duplicated or swapped phonemes (Wu et al., 2022). Therefore, models that include augmentations that alter the form of words more strongly might require higher temperatures to account for the greater variation in the resulting representations. This interaction of parameters makes the model difficult to tune. While Gao et al. (2022) find that dropout alone is sufficient for learning sentence representations, Chen et al. (2020) emphasize that a mixture of augmentations is best for learning image representations. The same seems to hold for learning single-word representations, as the dropout-only augmentation (see Table 12) considerably decreases the performance of the contrastive model.

This is also one of the clearer limitations of the presented method. Beyond the fact that hyperparameters were not exhaustively tuned and augmentation probabilities were chosen heuristically, the current approach also uses a fixed global weighting between PMI and cosine similarity. A simple grid search for optimal per-language weights and thresholds on the WOLD dataset with English, Thai and Japanese yields substantially better results (F1 0.849 vs. 0.782). This suggests considerable room for improvement if near-optimal weights can be derived automatically, although it remains unclear how such weighting could be estimated without supervision. Another limitation, shared with most comparable methods (cf. List, 2024), is that the evaluation procedure cannot infer borrowing direction. Automatic threshold detection also depends on the similarity distribution containing an identifiable bi-modality, which may not hold for all language pairs. And as with any method based purely on string comparison, deeper borrowing relations where there has been a complete adaptation to the recipient language remain undetectable.

Future work may move in several directions: a more systematic exploration of semantic similarity, testing alternative or stronger contrastive frameworks that better handle variability in word length, and developing richer, linguistically informed augmentations. Additionally, the accompanying command-line tool enables exploratory analysis of contact zones across large multilingual datasets. Finally, appropriately adjusted, the same framework could also be adapted to the conceptually related task of automated cognate detection.

Overall, the results indicate that self-supervised similarity learning offers a viable path toward scalable and language-agnostic borrowing detection. While many components of the approach have potential for improvement, the framework presented here offers a practical, scalable tool that can support researchers in identifying potential borrowings across a wide range of datasets.

## Acknowledgements

This research was supported by the DFG Centre for Advanced Studies in the Humanities Words, Bones, Genes, Tools (DFG-KFG 2237).

## Data and code availability

The data and code for the experiments in this paper, along with tools for running additional experiments, are available at https://github.com/TGH-2020/CWE_BorDetect and on Zenodo (https://doi.org/10.5281/zenodo.17778855).